\pdfoutput=1
\relax
\documentclass[letterpaper]{article} 
\usepackage[utf8]{inputenc}
\usepackage{aaai20}  
\usepackage{times}  
\usepackage{helvet} 
\usepackage{courier}  
\usepackage[hyphens]{url}  
\usepackage{graphicx} 
\usepackage{url} 
\usepackage{tikz}
\usepackage{amsfonts}       
\usepackage{amsmath}
\usepackage{amssymb}
\usepackage[ruled]{algorithm2e}

\SetKwIF{If}{ElseIf}{Else}{if}{}{else if}{else}{end if}%
\SetKwFor{While}{while}{}{end while}%
\SetKwRepeat{Do}{do}{while}

\SetKwInOut{Parameter}{Hyper-parameter}
\usepackage{amsmath}
\DeclareMathOperator*{\argmax}{arg\,max}
\DeclareMathOperator*{\argmin}{arg\,min}

\usepackage{graphicx}  
\DeclareMathOperator*{\Exp}{\mathbb{E}}
\title{Neural Architecture Search for Class-incremental Learning}
\author{ Shenyang Huang\textsuperscript{1}, Vincent François-Lavet\textsuperscript{1}, Guillaume Rabusseau\thanks{Canada CIFAR AI chair}\textsuperscript{2}\\
    \footnotesize shenyang.huang@mail.mcgill.ca,  vincent.françois-lavet@mcgill.ca, guillaume.rabusseau@umontreal.ca\\
    Mila, McGill University, Montreal, Canada\textsuperscript{1} \\
    DIRO - Mila, Universit\'{e} de Montr\'{e}al, Montreal, Canada\textsuperscript{2}}

\begin{document}

\maketitle

\begin{abstract}
In class-incremental learning, a model learns continuously from a sequential data stream in which new classes occur. Existing methods often rely on static architectures that are manually crafted. These methods can be prone to capacity saturation because a neural network's ability to generalize to new concepts is limited by its fixed capacity. To understand how to expand a continual learner, we focus on the neural architecture design problem in the context of class-incremental learning: at each time step, the learner must optimize its performance on all classes observed so far by selecting the most competitive neural architecture. To tackle this problem, we propose Continual Neural Architecture Search~(CNAS): an autoML approach that takes advantage of the sequential nature of class-incremental learning to efficiently and adaptively identify strong architectures in a continual learning setting. We employ a task network to perform the classification task and a reinforcement learning agent as the meta-controller for architecture search. In addition, we apply network transformations to transfer weights from previous learning step and to reduce the size of the architecture search space, thus saving a large amount of computational resources. We evaluate CNAS on the CIFAR-100 dataset under varied incremental learning scenarios with limited computational power~(1 GPU). Experimental results demonstrate that CNAS outperforms architectures that are optimized for the entire dataset. In addition, CNAS is at least an order of magnitude more efficient than naively using existing autoML methods.


\end{abstract}

\section{Introduction}


Continual learning, or lifelong learning~\cite{Review} is the ability to acquire new knowledge while retaining previously learned experiences, and  is one of the modern challenges of artificial intelligence. Various methods have been proposed to tackle continual learning~(referred to as continual learners). As seen in Table~\ref{table:comparison}, some continual learners rely on a static architecture which is manually crafted. These methods are susceptible to the phenomenon of~\emph{capacity saturation}~\cite{sodhani2018training}, where a neural network's ability to generalize to new concepts is limited by its fixed capacity. 


\begin{table*}[t]
    \centering 
    \caption{Approaches used by some recent methods to various challenges in continual learning.}\smallskip 
    \resizebox{2.0\columnwidth}{!}{
    \smallskip\begin{tabular}{l | l | l | l | l | l} 
    Challenge & CNAS & RCL~\cite{RCL} & DEN~\cite{DEN} & A-GEM~\cite{A-GEM} & iCaRL~\cite{iCaRL} \\
    \hline
    Catastrophic forgetting & knowledge rehearsal & weight freezing & network split & episodic memory & rehearsal prototype \\
    Capacity saturation & grow width, depth & grow width & grow width & no growth & no growth \\
    Rely on task descriptors & no & yes & yes & yes & no \\
    \end{tabular}
    }
    \label{table:comparison} 
\end{table*}

In this paper, we propose to continuously adapt the neural network architecture as new data arrive and we focus on the class-incremental learning setting. Rebuff~et~al.~\cite{iCaRL} introduced \emph{class-incremental learning} where an algorithm learns continuously from a sequential data stream in which new classes occur. Motivated by designing an expandable continual learner, we aim to solve the \emph{continual architecture design} problem where at each time step of class-incremental learning, the learner must optimize its performance on all observed classes so far by selecting the most competitive neural architecture. 

Any continual learner faces the challenge of \emph{catastrophic forgetting} where learning new information interferes with previously acquired knowledge~\cite{forgetting}, that is, the learner forgets how to perform old tasks when new ones are learned. In the continual learning literature, a constraint on the storage of past data is often enforced such as in~\cite{A-GEM} and~\cite{iCaRL}, and preventing catastrophic forgetting is thus one of the main focus of existing approaches to continual learning. In contrast, we store all data that has been observed in the past and maintain a growing dataset as more training examples become available. Catastrophic forgetting is then addressed by rehearsing on past data. We argue that this is a realistic setting since data storage is rarely an issue when compared to computation time. This allows us to fully use the available data to optimize the architecture selection, while still being computationally efficient. 

Recent techniques for automatically designing deep neural networks using reinforcement learning~(RL) agents have shown promising results. Methods such as Neural Architecture Search~(NAS)~\cite{NAS} and Efficient Architecture Search~(EAS)~\cite{EAS} employ a policy gradient approach called REINFORCE~\cite{REINFORCE}, allowing for high flexibility in the policy network design. EAS further proposes to use Net2Net~\cite{Net2Net} transformations to initialize sampled architectures, thus achieving  huge computational savings.

To address the continual architecture design problem, we propose Continual Neural Architecture Search~(CNAS). CNAS consists of three parts: a task network for solving the classification task, a deep reinforcement learning based meta-controller for adaptively exploring the architecture search space and a heuristic function for deciding when to expand the continual learner. Each time new data arrive, the meta-controller generates candidate architectures using Net2Net~\cite{Net2Net} transformations of the current task network. The decision of whether to expand the current architecture is based on a heuristic function of the performance of all the candidate architectures on a held-out dataset. This process allows the network structure to adaptively evolve in reaction to arrival of new classes or to other changes in the data distribution. 

The autonomous nature of CNAS makes it an autoML approach~\cite{autoML,RobustautoML}, offering an efficient and off-the-shelf learning system that avoids the tedious tasks of manually selecting the correct neural architecture at each time step. As the observed dataset becomes more complex or includes examples from multiple training distributions, manually designing the architecture for a continual learner is not only time-consuming but also increasingly difficult. Therefore, reducing human intervention is a natural progression to develop robust and self-sufficient continual learners. 
\paragraph{Summary of the contributions} We formalize the continual architecture design problem and experimentally show that dynamically adjusting the neural architecture of a continual learner results in stronger performance than using a static architecture. To the best of our knowledge, our proposed method CNAS is the first approach for the continual architecture design problem in class-incremental learning, as well as the first continual autoML method. Our experiments on the CIFAR-100 dataset~\cite{CIFAR} shows that CNAS constitutes a sound and promising approach to various class-incremental learning scenarios. In particular, \emph{CNAS automatically designs parameter-efficient networks that outperforms those optimized for the entire CIFAR-100 dataset} at each time step of the learning process. Furthermore, when compared to the naive approach of conducting a full-scale neural architecture search at each time step, \emph{CNAS is at least an order of magnitude faster than naively using alternative autoML methods}. 

\section{Preliminaries: Continual Learning} \label{Sec:continual-learning}

In this section, we introduce the continual learning setting. In particular, we formalize class-incremental learning and describe how it is different from the related setting of task-incremental learning. Then, we explain the continual architecture design problem in class-incremental learning. 
\subsection{Class-incremental Learning} \label{sub:class-incremental}
In the class-incremental learning setting, a model learns continuously from a sequential data stream in which new classes occur~\cite{iCaRL}. At any time step, the learner is required to perform multi-class classification for all classes observed so far. Formally, the goal of class-incremental learning is to learn, at each time step $T$, a classifier $f\colon\mathcal{X}\to\mathcal{Y}$ given the aggregation of the datasets seen up to now ($D_1,D_2,\cdots,D_T$), where each dataset \[ D_t = \{(x_{1},y_{1}),\cdots, (x_{n_t},y_{n_t})\}\subset \mathcal{X}\times\mathcal{Y}. \]
Here, $\mathcal X$ is the input space and $\mathcal Y \subseteq \mathbb{N}$ is the set of categories. 
At each time step $t$, new classes can be introduced into the training data. 
Denoting by $\mathcal Y_t = \{y_i\mid (x_i,y_i)\in D_t\}\subset \mathcal Y$ the set of classes present in $D_t$, we assume that each dataset $D_t$ is identically and independently drawn from the distribution $\mathcal D_{|Y\in\mathcal Y_t}$ where $\mathcal D$ is an unknown distribution over $\mathcal X \times \mathcal Y$ and $\mathcal D_{|Y\in \mathcal Y_t}$ denotes that $\mathcal D$ is conditioned on labels belonging to $\mathcal Y_t$. In this setting, the learning objective at time $t$ corresponds to identifying an hypothesis $f_t$ that minimizes the risk over the classes seen so far:
\begin{equation}\label{eq:1}
f_t = \argmin_{f} \: \Exp_{(X,Y) \sim \mathcal D_{|Y\in\mathcal Y_1\cup\cdots \cup \mathcal Y_t}}  \mathcal{L}(f(X), Y) \; ,
\end{equation}
where $\mathcal{L}$ is a loss function penalizing prediction errors over the random variables $(f(X),Y)$. The simplest scenario for class-incremental learning is the one where $k$ new classes are introduced at each time step which we refer to as \emph{$k$-class incremental learning}. This is the setting that most existing literature have experimented with. In this work, we also consider more realistic continual learning scenarios where (a)~not all training data for a particular class is available at once~(i.e., data for one class can be spread out over several distant time steps) and~(b) the number of unseen classes arriving at each time step is unknown.

In accordance with the learning objective defined in Eq.~\eqref{eq:1}, a natural metric to evaluate the performance of a model at test time is the \emph{average incremental accuracy} introduced in~\cite{iCaRL}. The average incremental accuracy at time step $t$ is the test accuracy of the model $f_t$ on the part of the test data consisting only of the classes seen up to time $t$:
\begin{equation}\label{eq:2}
    \mathbf{Average \; Incremental \; Accuracy} = \dfrac{1}{C} \sum_{i=1}^{C} R_i \; ,
\end{equation}
where $C = |\mathcal Y_1\cup\cdots \cup \mathcal Y_t|$ is the total number of classes seen until time $t$ and $R_i$ is the test  accuracy of the model $f_t$ on category $i$ discriminating from $C$ classes.
\begin{figure}[t]
    \centering
    \includegraphics[width=0.9\columnwidth]{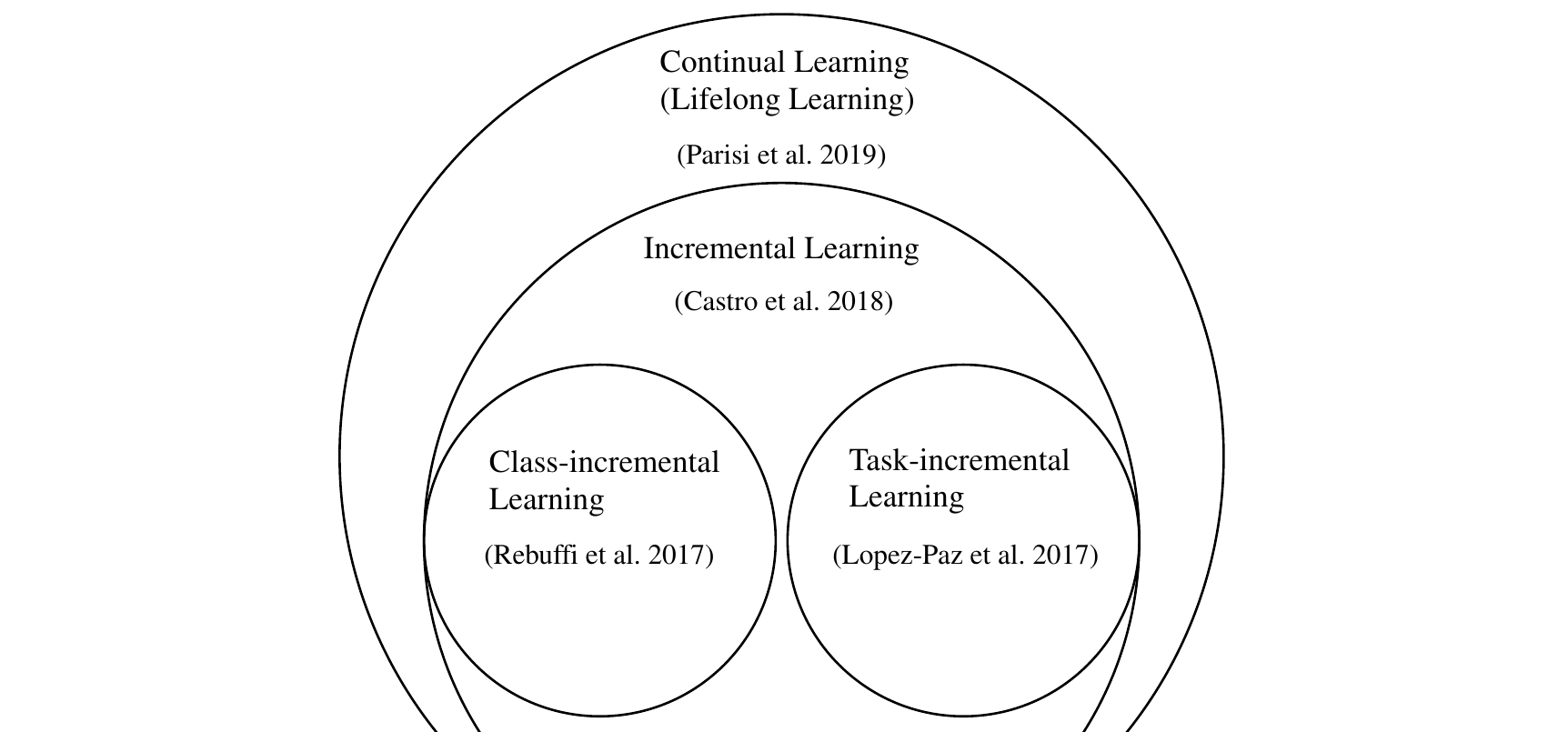} 
    \caption{The Venn diagram of continual learning with canonical references.}
    \label{Fig:CL}
\end{figure}

\subsection{Related Work and Task-incremental Learning} 

Lopez-Paz and Ranzato~\cite{GEM} defined the goal of continual learning as learning a predictor $f : \mathcal X \times  \mathcal T \rightarrow  \mathcal Y$ where $ \mathcal T$ refers to a set of task descriptors. Often in experiments~\cite{RCL,GEM,A-GEM}, an image classification dataset such as CIFAR100~\cite{CIFAR} or MNIST~\cite{MNIST} is separated into $N$ tasks~(each containing $k$ categories). Therefore, the predictor becomes dependent on the task descriptor to first identify which subset of categories the sample belongs to, before performing $k$-nary classification within the given subcategories. Because a task descriptor $t_i$ has to be given with each feature vector $x_i$, we consider the related continual learning definition proposed by Lopez-Paz and Ranzato~\cite{GEM} as \emph{task-incremental learning}~(using the same terminology as~\cite{van2018generative}), a separate learning paradigm from class-incremental learning. An illustration of the different settings in continual learning can be seen in Figure~\ref{Fig:CL}. In another work on task-incremental learning, Xu~et~al.~\cite{RCL} use a reinforcement learning agent to decide how many nodes or filters to add to the  layers of a fixed depth neural network. Since CNAS allows to increase both the width and depth of a neural network, it explores a more complex architecture space. 

\subsection{Continual Architecture Design} \label{sub:design}

In this work, we define \emph{continual architecture search} as the setting where, at each time step $t$, the continual learner must select the best neural architecture for classifying all classes seen so far. To tackle this setting,  we assume that the learner has access to all the data up to time $t$. We further impose a constraint with practical settings in mind: \emph{the initial architecture at $t=1$ is selected based on the initial dataset $D_1$ only}. Continual architecture search is concerned with hyperparameter optimization on a growing dataset while architecture search is traditionally conducted on a fixed training distribution. This difference implies that the architecture search space is continually growing thus making exhaustive search methods~(such as grid search) intractable from a computational standpoint. In contrast, CNAS takes advantage of the sequential nature of class-incremental learning by (i)~limiting the architecture search space by considering the structure of the task network from the previous step as a starting point and (ii)~using Net2Net techniques to rapidly transfer weights from previous step, 

\section{Continual Neural Architecture Search~(CNAS)}

In this section, we present our proposed method: Continual Neural Architecture Search~(CNAS). At any given time step $t$, CNAS provides a deep neural network with trained weights that is able to classify all observed categories so far. There are three components: a \emph{task network}, a \emph{meta-controller} and a \emph{heuristic function}.

The task network performs classification for all observed classes and is implemented as a standard deep neural network with convolution (CNN)~\cite{CNN}, maxpooling, dropout~\cite{dropout} and fully-connected layers. At each time step, the number of neurons in the last layer of the task network is equal to the number of observed classes $C$, and through the softmax activation function, each output neuron predicts the conditional probability of a category given the input. In class-incremental learning, new neurons are added to the output layer each time new categories appear~(these neurons are initialized with a zero-mean normal distribution for the weight matrix and zero for the bias term). 

The meta-controller is specialized in generating an architecture search policy to sample new candidate architectures for the task network when new classes arrive. The controller is implemented as a deep reinforcement learning agent. The role of the meta-controller is only to guide the architecture sampling process, by selecting promising architectures to try out based on experiences gathered from previous time steps. The selection of the best architecture out of the sampled ones is based on a validation set. This can be seen as a one-step ahead planning guided by the meta-controller to explore good candidates. 

Lastly, the heuristic function considers the validation performance of all the sampled architectures and decides if an expansion is beneficial in the current step. Preventing unnecessary expansions will reduce the computational time in subsequent steps as well as increasing the parameter efficiency of the task network. 

\subsection{Training Procedure}

\begin{algorithm}[t]
    \caption{CNAS IncrementalLearn}
    \SetAlgoLined
    \KwIn{Past dataset $D_{|Y\in\mathcal Y_1\cup\cdots \cup \mathcal Y_{t-1}}$. New dataset $D_{|Y\in\mathcal Y_t}$. Task network $\beta_{t-1}$.}
    \KwOut{New task network $\beta_{t}$}
    \BlankLine
    Concatenate $D_{|Y\in\mathcal Y_1\cup\cdots \cup \mathcal Y_{t-1}}$ with $D_{|Y\in\mathcal Y_t}$ \;
    Expand output dimension for $\beta_{t-1}$ \;
    $(\beta_{t-1},v_{t-1})\leftarrow$ Train $\beta_{t-1}$ with $D_{|Y\in\mathcal Y_1\cup\cdots \cup \mathcal Y_{t}}$ \;
    $(\beta_{*},V_{sampled})\leftarrow$ ArchSearch~$(\beta_{t-1}$, $D_{|Y\in\mathcal Y_1\cup\cdots \cup \mathcal Y_{t}})$ \;
    ($\beta_t)\leftarrow$ HeuristicFunc~$(\beta_{t-1},\beta_{*},v_{t-1},V_{sampled})$ \;
    Train $\beta_{t}$ with $D_{|Y\in\mathcal Y_1\cup\cdots \cup \mathcal Y_{t}}$ \;
    Return $\beta_{t}$ \;
    \label{Algo:1}
\end{algorithm}

Algorithm~\ref{Algo:1} describes the training procedure for CNAS when a new dataset arrives. The task network is first trained with a combined dataset of past and new examples and is then used as the starting point for \textit{ArchSearch}~(Algorithm~\ref{Algo:2}). \textit{ArchSearch} then outputs the validation accuracies of all the sampled architectures and the best performing candidate architecture. \textit{HeuristicFunc}~(Algorithm~\ref{Algo:3}) then decides if expanding the current task network is beneficial based on the validation performance differences between the sampled candidate architectures and the existing architecture. When deciding to expand, the best performing sampled architecture becomes the new task network structure. If no expansion is needed, no change is made to the current architecture. This new task network is then further trained on the available data to ensure it has converged. The number of candidate architectures that can be sampled per time step is a hyper-parameter of our algorithm and it controls the trade-off between computational complexity and exploration depth.

\begin{algorithm}[t]
    \SetAlgoLined
    \Parameter{Sample size $n$. Epoch limit $l$.}
    \KwIn{Dataset $D_{|Y\in\mathcal Y_1\cup\cdots \cup \mathcal Y_{t}}$. Current architecture $\beta_{t-1}$.}
    \KwOut{Best performing architecture $\beta_{*}$. List of validation accuracy of sampled architectures $V_{sampled}$.}
    \BlankLine
    \For{$i = 1,\dots, n$}{
        Generate candidate architecture $\hat{\beta}^i$ from RL agent \;
        Initialize $\hat{\beta}^i$ by weights from $\beta_{t-1}$ using Net2Net\;
        Train $\hat{\beta}^i$ for $l$ epochs with $D_{|Y\in\mathcal Y_1\cup\cdots \cup \mathcal Y_{t}}$\;
        Append validation accuracy $\hat{v}_i$ of $\hat{\beta}^i$ to $V_{sampled}$ \;
    }
    Update RL agent with all $(\hat{\beta}^i, \hat{v}_i)$ pairs\;
     \textit{// Return the candidate architecture with best accuracy:}\\
    Let $\beta_* = \hat{\beta}^j$ where $j=\argmax_{i=1,\dots,n}\{\hat{v}_i\}$  \;
    Return ($\beta_{*}$, $V_{sampled}$) \;
    \caption{CNAS ArchSearch}
    \label{Algo:2}
\end{algorithm}

One could greedily expand the continual learner at each time step~(i.e. always set $\beta_t$ to $\beta_*$ in Algorithm~\ref{Algo:1}). However, this can not only reduce parameter efficiency but also potentially affect future performance. The heuristic function~(\textit{HeuristicFunc}, see Algorithm~\ref{Algo:3}) is designed to evaluate the benefit of expansion based on the difference in validation performance between all sampled architectures and the existing architecture. If capacity saturation occurs, expanding the architecture will likely result in performance improvement and architecture expansion is considered necessary. However, when only a small portion of expanded structures shows gains in performance then it is likely that these improvements are due to the randomness in network training and architecture expansion is not required. 

\begin{algorithm}[t]
    \caption{CNAS HeuristicFunc}
    \SetAlgoLined
    \KwIn{Current architecture $\beta_{t-1}$. Best performing sampled architecture $\beta_{*}$. Performance of current architecture $v_{t-1}$. Performance of sampled architectures $V_{sampled}$.}
    \KwOut{New task network architecture $\beta_t$}
    \BlankLine
    \textit{// Calculate number of negative improvements:} 
    $N_{negative} = |\{ v \in V_{sampled} \mid v < v_{t-1}\}| $\;
    \uIf{$N_{negative} < \frac{|V_{sampled}|}{2}$ and $\mathrm{mean}(V_{sampled}) > v_{t-1}$}{
         Return $\beta_{*}$ ; $//$ Architecture is expanded.
     }
     \Else{
         Return $\beta_{t-1}$ ; $//$ No expansion.
     }
    \label{Algo:3}
\end{algorithm}

\subsection{Net2Net Transformations}
To save the computational cost of training each sampled architecture from scratch, we use a transfer learning technique called Net2Net~\cite{Net2Net}. Net2Net enables a rapid transfer of information from one neural network to another by expanding/creating fully-connected and convolutional layers using two types of operations. Net2WiderNet operations replace a given layer by a wider one~(more units for fully-connected layers or more filters for convolutional layers) while preserving the function computed by the network. Net2DeeperNet operations insert a new layer that is initialized as an identity mapping between two existing layers, thus preserving the function computed by the neural network. More formally, Net2DeeperNet replaces a layer $h^{(i)} = \phi(W^{(i)}h^{(i-1)})$ with two layers $h^{(i)} = \phi(I \phi(W^{(i)}h^{(i-1)}))$ where $I$ is the identity matrix. However, the last equality is true only  if the activation function $\phi$ is such that $ \phi(I\phi(v)) = \phi(v)$ for all vectors $v$, which holds for the rectified linear activation~(ReLU). Therefore, we use ReLu activation for all hidden layers. 

Net2WiderNet and Net2DeeperNet operations can be applied sequentially to grow the original network in both width and depth. In this way, any architecture that is strictly larger than the original can be initialized to preserve the function computed by the original network. This allows CNAS to use a trained network as a starting point for architecture search and quickly initialize new larger architectures. By using Net2Net, the capacity of the task network can be expanded efficiently and dynamically for stronger performance as new data become available. Further details regarding Net2Net transformations are provided in the original paper~\cite{Net2Net}.

\section{Reinforcement Learning Agent}

We use the policy gradient method REINFORCE~\cite{REINFORCE} and design two independent policy networks for taking Net2WiderNet actions and Net2DeeperNet actions respectively, with the simplifying assumption that they are independent.

We describe continual architecture design as an RL problem: at each step, an agent observes the current state $s_t$ of the environment and samples actions (=network transformations) $a_t$ according to a stochastic policy $\pi(a_t|s_t)$. For each sampled action, it observes a reward signal $r_t$, which is used along with a step size $\alpha$ to improve the policy for future time steps. For computational efficiency, a fixed number of architectures is sampled at each time step using the RL agent. The planning horizon is limited to one time step. 
Limiting the horizon acts as a complexity control method~\cite{jiang2015dependence} and results in only optimizing for the current distribution. The REINFORCE algorithm is simplified to:
\begin{equation}\label{eq:3}
\theta \leftarrow \theta + \alpha r_t \nabla_\theta \ln \pi(a_t|s_t,\theta)
\end{equation}
where $\theta$ represents the parameters of the policy networks for Net2WiderNet and Net2DeeperNet.

The architecture search for each time step is summarized in Algorithm~\ref{Algo:2}. Any sampled architecture is trained for at most $l$ epochs using early stopping. Due to the benefit of weight transfer through Net2Net transformations, sampled architectures only require training for a low number of epochs in practice.  



\subsection{Policy Networks}

\begin{figure}[t]
    \centering
    \includegraphics[width=0.9\columnwidth]{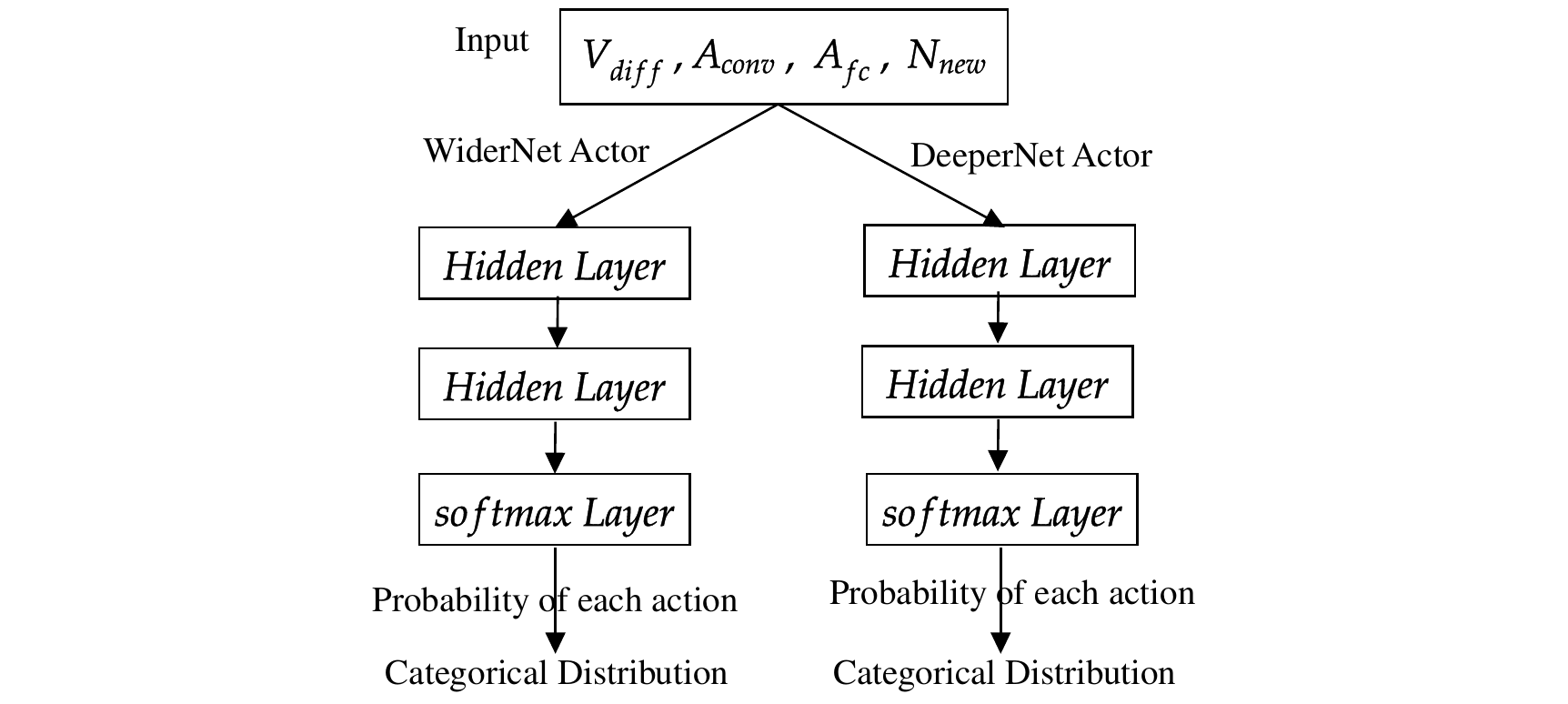}
    \caption{Flow chart of the policy network}
    \label{Fig:RL_flow}
\end{figure}

The policy networks for Net2WiderNet and Net2DeeperNet, referred to as wider actor and deeper actor respectively, are identical in design as seen in Figure~\ref{Fig:RL_flow}, but trained independently. Encoding the task network's architecture in details into the state $s_t$ might be of little use to the RL agent as such states are almost never repeated~(since the architecture is continuously expanding). Therefore, we only include the number of convolutional layers and the number of fully-connected layers of the task network in $s_t$~(denoted by $A_{conv}$ and $A_{fc}$ respectively). Moreover, to measure the disparity between the current training distribution $D_t$ and the previous one $D_{t-1}$, the difference in validation accuracy of the task network on these two distributions is included in the state space~(denoted by $V_{diff}$). Lastly, the number of new classes received by the continual learner at the current time step is also added~(denoted by $N_{new}$). 

The wider and deeper actors decide the number of Net2WiderNet and Net2DeeperNet transformations to take respectively, and are implemented as multilayer perceptrons. Both the input and hidden layers have ReLU activation while the output layer of the actor networks has a softmax activation. The $i$-th output neuron corresponds to the probability of taking $(i-1)$ transformations and the first neuron always represents not taking any transformations. The predicted probability is then used as input to a categorical distribution out of which actions are selected. In this way, with the same input state space, the number of transformations selected by the actor networks is stochastic. 

\subsection{Reward Design}
To best decide the number of transformations needed for each time step, we design a reward function based on the performance of the newly transformed architecture, compared to the existing one~(measured with average incremental accuracy from Equation~\ref{eq:2}). We consider the difference in validation accuracy between the original architecture and the sampled architecture,
$r_t = v' - v_{t-1}$.
Here $r_t$ is the reward signal given to the agent at time step $t$ after deciding on the number of Net2Net transformations while $v'$ and $v_{t-1}$ respectively stands for the validation accuracy of the sampled architecture and of the original architecture on the current dataset after training. Therefore, any architecture that performs worse than the original will provide a negative reward signal while a better architecture will yield positive one. To obtain better reward signals for learning, the rewards are normalized into [-1,1] range within all architectures sampled at time $t$. Lastly, we add an entropy term to the reward function in order to improve policy optimization~\cite{ahmed2019understanding}. 

\section{Experiments}
We now describe our experimental setup and details regarding implementation of CNAS~(the code will be made publicly available). We repeat each experiment three times with different random seeds and report the standard deviation with error bars.

\textbf{Dataset}  We split both the training set and test set of CIFAR-100 by class labels. The CIFAR-100 dataset contains a total of 60,000 images across 100 classes. In our experiments, each class is further split into 450 images as training set, 50 images as validation set and 100 images as test set. When a new class is introduced, all corresponding test data will start to be used for the calculation of average incremental accuracy. In the k-class incremental learning scenario, all corresponding training examples are presented as a new class is introduced. We also test scenarios where only a fraction of all training examples of a certain class becomes available at a time step. The examples contained in the validation set are used for architecture selection. The arrival order of the classes is based on the default labels given by the CIFAR-100 dataset. Each experiment starts with some initial classes~(known as the base knowledge and considered as the dataset for time step $0$). 

\textbf{Baselines} We compare CNAS with the following baselines:  (1) \emph{SA}~(Static Architecture): a continual learner with a static architecture that is selected given the knowledge of all 100 classes at once~(i.e., optimized on the entire CIFAR-100 dataset); (2) \emph{RAS}~(Random Architecture Search): a continual learner that greedily expands its architecture whenever the best sampled architecture has a stronger validation performance. It uses a uniformly random architecture sampling strategy; (3) \emph{RAS-HF}~(Random Architecture Search with Heuristic Function): random architecture search with the same heuristics function as CNAS~(see Algorithm~\ref{Algo:3}). RAS and RAS-HF are compared with CNAS in the ablation study. We use average incremental accuracy on the test set as the evaluation metric. 

\textbf{Implementation} We implement CNAS with Keras~\cite{Keras} using Tensorflow~\cite{Tensorflow} as the backend framework. All approaches are trained using the ADAM~\cite{ADAM} optimizer with a learning rate of $10^{-4}$ and other parameters set to default values. All training is conducted with mini-batches of size 128. The task network is trained until convergence~(with early stopping) both before and after the architecture search at each time step. Both the wider and deeper actors are implemented as a multilayer perceptron with 2 hidden layers, each having 128 neurons. The learning rate of the RL agent is 0.001 and the entropy regularization term is scaled by a factor of 0.01.

\paragraph{K-class Incremental Learning} We compare the performances of CNAS and SA on $k$-class incremental learning experiments on CIFAR-100 for $k=2$ and $k=10$. For 2-class incremental learning, CNAS samples 20 architectures at each time step and can take at most 3 Net2WiderNet and 3 Net2DeeperNet actions. For 10-class incremental learning, 50 architectures are sampled at each time step and at most 10 Net2WiderNet and 5 Net2DeeperNet transformations can be taken. The base knowledge is the first 10 classes and the initial architecture for CNAS is optimized for the base knowledge only. 

\begin{figure}[t]
\centering
\includegraphics[width=0.9\columnwidth]{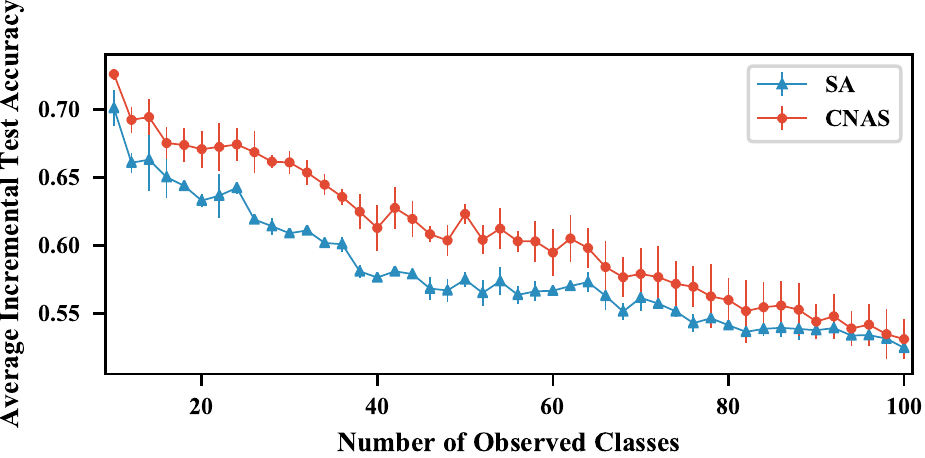}
\caption{Performance of SA and CNAS in 2-class incremental learning}
\label{Fig:2-class}
\end{figure}

\begin{figure}[t]
\centering
\includegraphics[width=0.9\columnwidth]{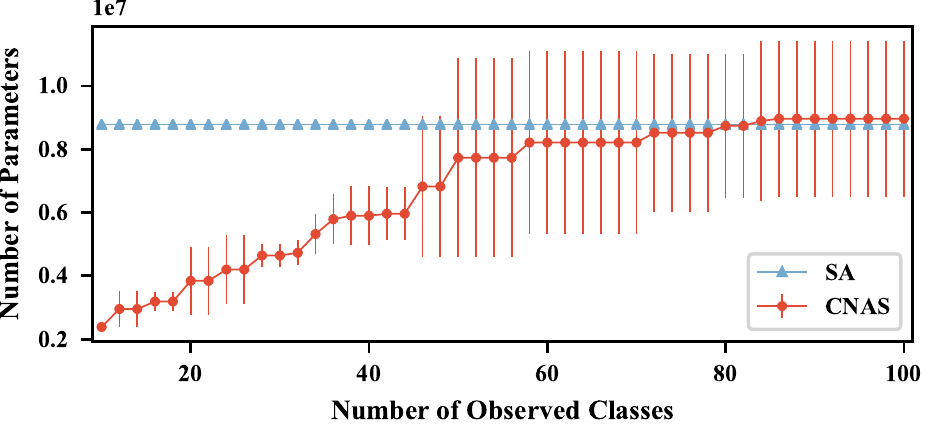}
\caption{Parameter Growth Curve in 2-class incremental learning experiment}
\label{Fig:2-class-params}
\end{figure}

\begin{figure}[t]
\centering
\includegraphics[width=0.9\columnwidth]{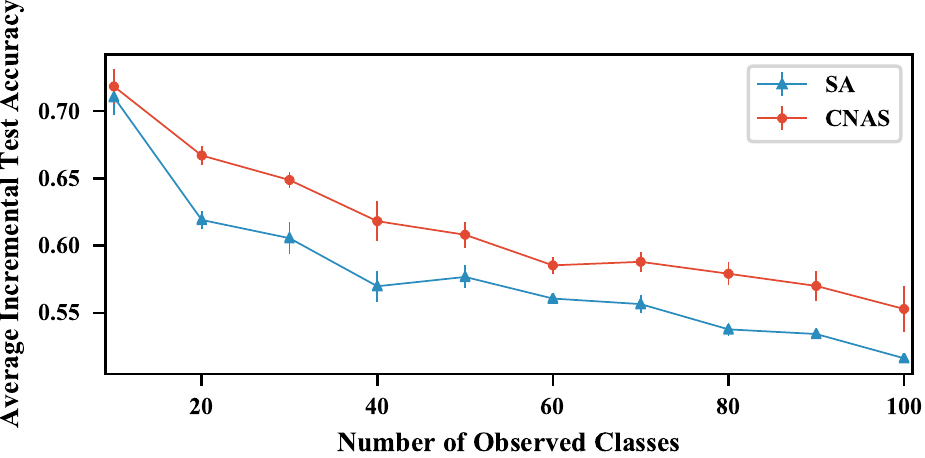}
\caption{Performance of SA and CNAS in 10-class incremental learning}
\label{Fig:10-class}
\end{figure}

\begin{figure}[t]
\centering
\includegraphics[width=0.9\columnwidth]{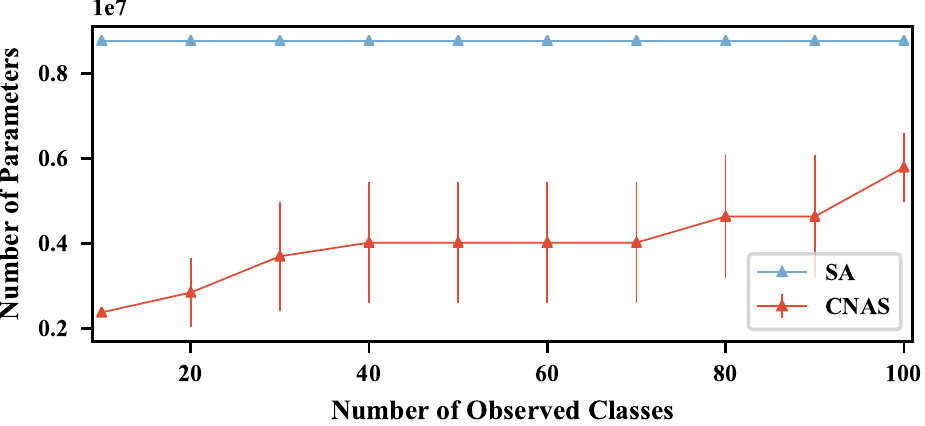}
\caption{Parameter Growth Curve in 10-class incremental learning experiment}
\label{Fig:10-class-params}
\end{figure}

From Figure~\ref{Fig:2-class} and~\ref{Fig:10-class}, we can see that CNAS outperforms SA in terms of average incremental test accuracy. Furthermore, CNAS consistently uses less parameters than SA, as shown in Figure~\ref{Fig:2-class-params} and~\ref{Fig:10-class-params}. Due to having a smaller initial structure, CNAS is able to generalize better than SA on the base knowledge. As more classes are introduced, the difficulty of the task increases and larger architectures are required to avoid capacity saturation. Note that there are many steps where CNAS chooses not to expand and maintains its architecture  (see Figure~s\ref{Fig:2-class-params} and~\ref{Fig:10-class-params}). The heuristics function ensures that only necessary expansions are taken. 

\begin{figure}[t]
\centering
\includegraphics[width=0.9\columnwidth]{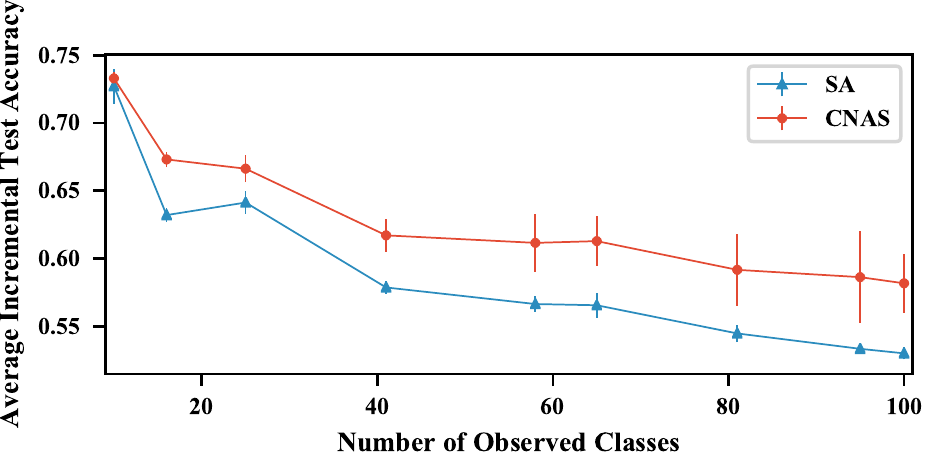}
\caption{Performance of SA and CNAS in mixed-class incremental learning}
\label{Fig:mixed-class}
\end{figure}

\begin{figure}[t]
\centering
\includegraphics[width=0.9\columnwidth]{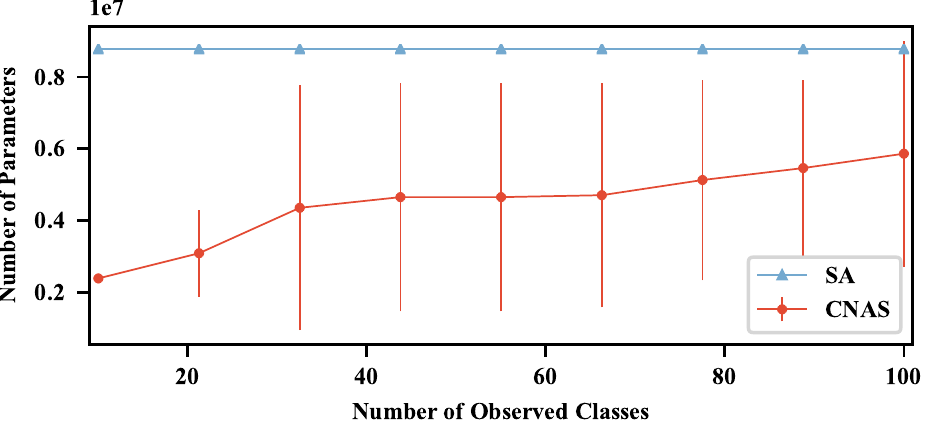}
\caption{Parameter Growth Curve in mixed-class incremental learning experiment}
\label{Fig:mixed-class-params}
\end{figure}

\paragraph{Mixed-class Incremental Learning} We introduce a more realistic incremental learning setting where the number of new classes as can vary at each time step and additional training data from already seen classes can arrive at later time steps~(referred to as \textit{mixed-class incremental learning}). In this experiment, the continual learner will receive all the training data from $k$ unseen classes and a portion $p$ of the training data of either an existing class or an unseen class. At each step, $k$ is chosen randomly from range~[1,19] and $p$ can be either 0.25 or 0.5. This scenario is motivated by the use case where the number of classes introduced at each step is unknown and data from some classes are spread-out over different time steps. CNAS can sample up to 30 architectures at each step and take a maximum of 5 Net2WiderNet and 5 Net2DeeperNet operations. We see in Figures~\ref{Fig:mixed-class} and~\ref{Fig:mixed-class-params} that CNAS significantly outperforms SA while using less parameters. This is because CNAS can identify the optimal architecture for the current training distribution and adapt its architecture accordingly.

\begin{figure}[t]
\centering
\includegraphics[width=0.9\columnwidth]{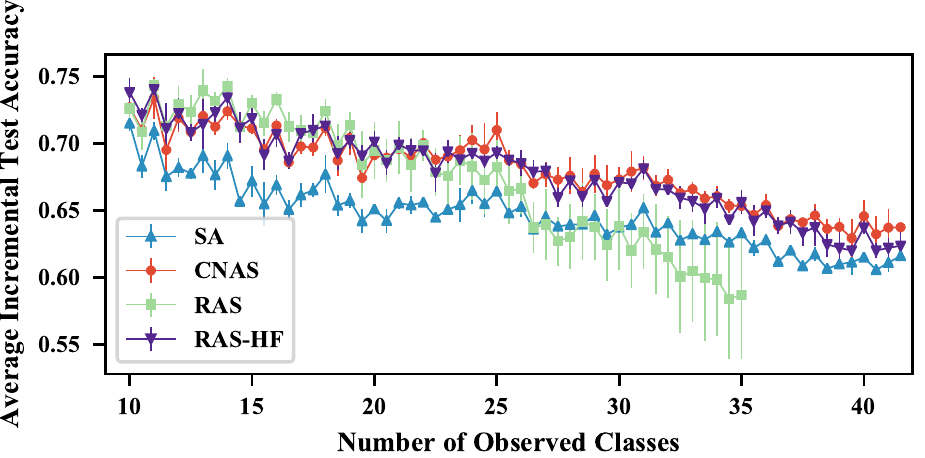}
\caption{Performance of SA, RAS, RAS-HF and CNAS in the ablation study experiment}
\label{Fig:half-class}
\end{figure}

\begin{figure}[t]
\centering
\includegraphics[width=0.9\columnwidth]{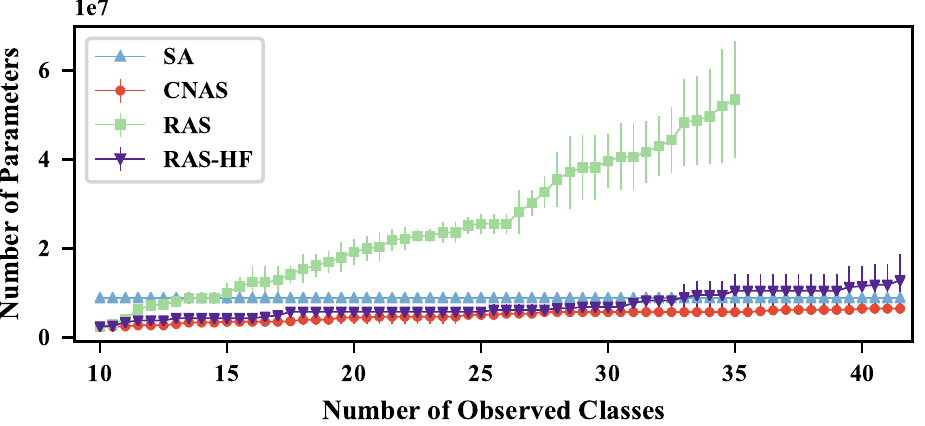}
\caption{Parameter Growth Curve in the ablation study experiment}
\label{Fig:half-class-params}
\end{figure}

\paragraph{Ablation Study} Lastly, we consider a new and difficult incremental learning scenario where only half of the training data of a class arrives at a time step. RAS contains no heuristic function nor the RL meta-controller when compared to CNAS. In comparison, RAS-HF has the heuristic function but lacks the meta-controller. In Figures~\ref{Fig:half-class} shows that CNAS has the best performance overall. In Figure~\ref{Fig:half-class-params}, we see that RAS greedily expands its architecture at the beginning and leads to complex models that are unable to generalize well to the task at hand. In addition, RAS leads to oversize models that become too costly to train using only 1 GPU~(this is why the curve ends earlier). This shows that the heuristic function is important to prevent over-expansions. The RL meta-controller is also important for CNAS as it learns to narrow down the architecture search space based on the current learning paradigm. In the first time steps, RAS-HF obtains performances similar to CNAS. This is because the RL agent requires experiences to adapt its policy from uniformly random to one that is tailored for the current incremental learning setting. When 36 classes are learned, CNAS starts to consistently outperform RAS-HF while having a smaller task network. Overall, both the heuristics function and the RL meta-controller are essential components of CNAS in class-incremental learning settings.

\paragraph{Computational Time} We report the average computational time on 1 GPU across 3 trials in this section. In 2-class incremental experiment, CNAS used 93 hours and explored 900 architectures. Note that CNAS can take advantage of multiple GPUs and train many sampled architectures in parallel. In comparison, early neural architecture search approaches such as NAS~\cite{NAS} performed architecture search on the CIFAR-10~\cite{CIFAR} dataset with 800 GPUs and trained 12,800 models from random initialization. A more recent approach such as EAS~\cite{EAS} also uses Net2Net techniques for weight transfer and they use 5 GPUs for 2 days to train 450 CNNs. In this regard, CNAS is at least one order of magnitude faster than naively using autoML alternatives such as EAS and NAS. 

In the ablation study experiment, CNAS used 26 hours and sampled 315 neural architectures. Each component in CNAS contributes to a greater computational efficiency. Without the RL meta-controller, RAS-HF also sampled 315 architectures but used 67 hours. Without the heuristic function, RAS explored 250 architectures while using 91 hours.

\section{Discussion}
CNAS requires very few hyper-parameters to be effective. The starting architecture of CNAS is optimized on the base knowledge~(training data at time step 0). The number of sampled architectures as well as the maximum number of Net2Net transformations should be selected based on the available computational resources. Given a larger search space and more sampled architectures, CNAS is likely to find stronger models. 


CNAS avoids greedily picking the architecture with the best validation performance, which can quickly lead to overparametrized models.
Indeed, the validation accuracy of any sampled architecture is not only determined by the effectiveness of the neural architecture but also affected by the stochasticity of parameter optimization. 
As seen in Figure~\ref{Fig:half-class-params}, the heuristic function in CNAS plays an important role to avoid unnecessary model expansions.



\section{Conclusion}
In this paper, we presented the problem of continual architecture design in class-incremental learning. We proposed CNAS, an efficient and economical autoML approach for continual learning. CNAS (i)~reuses trained weights through Net2Net, (ii)~implements an RL meta-controller to find the most effective architecture transformations and (iii)~uses a heuristic function to decide when to expand the current architecture. Various incremental learning experiments on the CIFAR-100 dataset show that CNAS consistently outperforms architectures that are optimized on the entire dataset. 

\subsection*{Acknnowledgments}
The authors gratefully acknowledge the support of the Natural Sciences and Engineering Research Council of Canada (NSERC) and the Canadian Institute for Advanced Research (CIFAR).


\bibliography{references}
\bibliographystyle{aaai}

\end{document}